\documentclass{article}
\usepackage[whole]{bxcjkjatype}

\usepackage{PRIMEarxiv}

\usepackage[utf8]{inputenc} 
\usepackage[T1]{fontenc}    
\usepackage{hyperref}       
\usepackage{url}            
\usepackage{booktabs}       
\usepackage{amsfonts}       
\usepackage{nicefrac}       
\usepackage{microtype}      
\usepackage{lipsum}
\usepackage{fancyhdr}       
\usepackage{graphicx}       
\usepackage{ascmac}
\usepackage{appendix}
\graphicspath{{media/}}     

\title{Technical Report: Small Language Model for Japanese Clinical and Medicine
}

\author{
  Shogo Watanabe \\
  National Cerebral and Cardiovascular Center Hospital, Japan \\
  Graduate School of Medicine and Faculty of Medicine, Kyoto University, Japan \\
  \texttt{watanabe.shogo@ncvc.go.jp} \\
}

\begin{document}
\maketitle

\begin{abstract}
This report presents a small language model (SLM) for Japanese clinical and medicine, named NCVC-slm-1. This 1B parameters model was trained using Japanese text classified to be of high-quality. Moreover, NCVC-slm-1 was augmented with respect to clinical and medicine content that includes the variety of diseases, drugs, and examinations. Using a carefully designed pre-processing, a specialized morphological analyzer and tokenizer, this small and light-weight model performed not only to generate text but also indicated the feasibility of understanding clinical and medicine text. In comparison to other large language models, a fine-tuning NCVC-slm-1 demonstrated the highest scores on 6 tasks of total 8 on JMED-LLM. According to this result, SLM indicated the feasibility of performing several downstream tasks in the field of clinical and medicine. Hopefully, NCVC-slm-1 will be contributed to develop and accelerate the field of clinical and medicine for a bright future.
\end{abstract}

\keywords{small language model (SLM) \and Japanese \and clinical and medicine}

\section{Introduction}
Large language model (LLM), based on Transformer \cite{Vaswani2017}, is widely spreading in many fields. Transformer-based language model is formulated by auto-regressive language model and implemented using causal masking on self-attention layer as follow:

\begin{equation}
  p(x) = \prod_{i=1}^{n} p(s_n | s_1, ..., s_{n-1}) \label{eq:arlm}
\end{equation}

In the context of language processing, the variables $s_i$ and $p(x)$ represent word tokens and probabilities, respectively.

The most famous Transformer-based language model, Generative Pre-training Transformer (GPT) \cite{gpt, gpt2, gpt3, gpt4} series, demonstrated surprising performances regardless of simply architecture and training manner. In particularly, ChatGPT ignited this main stream. Since Scaling laws \cite{scalinglaw2020, scalinglaw2022} was proposed, various LLMs have been developed and released in recent years \cite{llama,llama2,llama3,palm,palm2,falcon2023,gemma,gemma2,qwen,qwen2,claude3,cohere_for_ai_2024}. In the field of clinical and medicine, some LLMs were developed and released \cite{medalpaca, meditron, medicine-llm-13b, me-llama, medgemini, pfnet-medswallow}.
Gigantic LLMs achieve or exceed human-level ability on several benchmarks. Larger models show higher performance. However, they require high computational cost (even if using some quantization or light-weighting techniques).
Besides, huge LLMs have some disadvantages. Too large models increase inference time and slow response time, therefore they are low maneuverability and inconvenience to quickly try and error. In addition, they require high-end hardware to run. It is difficult to use large models in local environments. Therefore, client-server systems are an option to use large models, but they involve the risk of leaking personal or confidential information to a third-party server.

On the other hand, small language model (SLM) is proposed such as Phi series (phi-1, phi-1.5, and phi-3) \cite{phi1,phi1.5,phi3}, LLaMA 3.2 1B, gemma-2-2B, and so on. These SLMs indicate the feasibility of better performance and reduce computational costs. In particular, the paper of phi-1 proposed textbooks approach, which use only textbook quality data for training models. Focusing on a specific field and using textbooks approach, a well fine-tuned SLM suggested the potential for outperforming larger models in a particular domain. The motivation of this project is stimulated by those findings.

This paper notes a strategy for the development of SLM specialized to the field of clinical and medicine of Japanese text. A developed SLM is named NCVC-slm-1.

\section{Materials and Methods}

\subsection{Dataset}
NCVC-slm-1 was trained using two main corpus stimulated by textbooks approach.

\subsubsection{Common Corpus}
The first dataset is a widely used common corpus. NCVC-slm-1 used Japanese Wikipedia \cite{wiki40b} and OSCAR \cite{oscar} (OSCAR-2301). Following the policy that fewer parameter model and high-quality dataset, text corpus was focused on high-quality data only.
Japanese Wikipedia was used only text extracted from articles.
OSCAR is a large dataset based on Common Crawl. Although OSCAR is applied some filtering and removing duplication, web-based text datasets contain various topics documents, not proper contents, and noise. For screening corpus, some following strongly additional filtering were applied: filtering documents containing adult content words and copyright notices, screening documents containing not completing sentences, and removing duplication documents and sentences. These processes filtered approximately 68\% of data size.
Furthermore, text quality filtering based on transformer-based classifier were performed \cite{phi1}. First, 100,000 documents were randomly sampled from corpus and annotated text quality using Llama3-8B-Instruct (Llama-3-ELYZA-JP-8B-instruct). Annotation prompt format is あなたは誠実で優秀な日本人のアシスタントです。入力された文章が教科書レベルの品質かどうかを判断してください。回答は'高'または'低'のみを出力してください。(In English: You are an honest and excellent Japanese assistant. Determine whether the input text is of textbook quality. Please answer 'high' or 'low' only.).
Second, the last token embedding vector of the last hidden state was calculated on all documents. A random forest \cite{RF} classifier was trained using its embedding vector as features and annotated labels. A trained text quality classifier filtered documents that were considered low quality. Unfortunately, approximately only 1\% of data size was filtered in this process.

\subsubsection{Medicine Textbooks}
The second corpus is textbook quality dataset comprising clinical and medical knowledge scraped from the Web.
However, the amounts of scraping text focusing on clinical and medicine contents were only 49 MB. In order to augment the shortage of text about clinical and medicine, synthesized textbooks approach were performed.
Synthesized textbooks contain the topics of general and various diseases and drugs based on J-Medic and the list of items on the National Health Insurance drug price list announced on the website of Ministry of Health, Labour and Welfare, Japan. Their textbooks were 5 versions of previous language models (ELYZA-Llama2-7B-Fast-Instruct \cite{elyzallama2}, ELYZA-Llama3-8B-instruct \cite{elyzallama3}, StableLM-7B-Alpha-Instruct \cite{JapaneseStableLMInstructAlpha7Bv2}, StableLM-7B-Beta-Instruct, and Phi-3-mini-128K-instruct \cite{phi3}). These instructed models generated topics of 91,000 diseases and 9,000 drugs.
Moreover, ELYZA-Llama3-8B-instruct generated exercises text which similar to Japanese national medical licensing examinations and Japanese national pharmacists licensing examinations. Synthesized exercises dataset were prepared 5 samples per each topics.

Table \ref{tab:corpus} shows the overview of data size and number of tokens for pre-training corpus. Finally, the total number of unique tokens were approximately 9 billions.

\begin{table}[htbp]
  \centering
  \caption{The details of corpus for pre-training}
  \label{tab:corpus}
  \begin{tabular}{rrc}
   & \multicolumn{1}{c}{\textbf{Data Size}} & \textbf{Tokens} \\ \hline
  \multicolumn{1}{l}{\textbf{Common Corpus}} & & \\ \hline
  Wikipedia(Wiki-40B) & 2 GB & 0.55 B \\ \hline
  OSCAR(oscar2301) & 30 GB & 8.20 B \\ \hline
  \multicolumn{1}{l}{\textbf{Medicine Textbooks}} & & \\ \hline
  Scraping Medicine Textbooks & 49 MB & 0.01 B \\ \hline
  Synthetic Medicine Textbooks & 895 MB & 0.23 B \\ \hline
   & & 8.99 B
  \end{tabular}
\end{table}

\begin{figure}[htbp]
  \centering
  \includegraphics[scale=0.3]{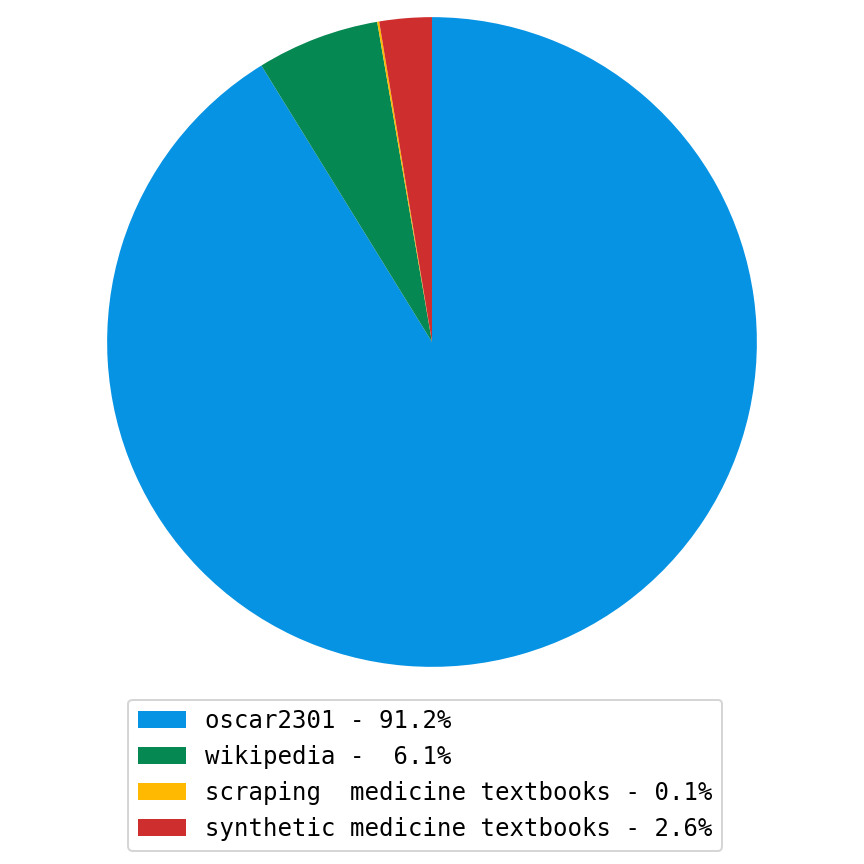}
  \caption{Data proportions in pre-training corpus}
  \label{fig:data-proportions}
\end{figure}

Figure \ref{fig:data-proportions} illustrates the pie chart of data proportions for pre-training. OSCAR occupies a large percentage of the whole corpus. Unfortunately, the augmented clinical and medicine textbooks were only 2.6\%.

\subsection{Tokenizer}
Generally speaking, refined pre-processing is considered an important process to enhance accuracy or performance. Huge LLMs may cover this process. However, given the scalability of NCVC-slm-1, proper pre-processing is considered to be necessary.
Pre-processing is performed in 2 steps text cleaning and morphological analysis before tokenization.

First, text cleaning is applied to raw text. This process contains replacing specific characters, normalizing Unicode and spelling inconsistencies, removing personal information and web-specific notations, unifying symbols, reducing redundancy, and so on.
Japanese characters do not only have full-width and half-width character sets but also are used mixing punctuations frequently. Here, full-width period and comma are replaced with full-width 。 and 、, respectively. This is because their symbols are normalized to half-width symbols by subsequent Unicode normalization.
In NCVC-slm-1, Unicode normalization is used Normalization Form Compatibility Composition (NFKC). Furthermore, it is normalized spelling inconsistencies such as hyphens, chōonpus, and tildes. Restoration is applied continuous three dots to ellipsis(…) and °C to ℃.
In the perspective of protecting privacy, it is applied removing personal information such as phone numbers and e-mail addresses using regex processing. In addition, web-specific notations are removed such as URL, @ account names, and \# tags.
Symbols include similar meaning are unified because of reducing the wastes of vocabularies.
After these cleanings, other than the using characters are deleted because of limiting the space of characters.
In order to eliminate superfluous repetitions, the unnecessary continuous characters are removed.
Finally, white-space are replaced with meta-space (U+2581) because white-space is used in the pre-processing of subword tokenizer.

Second, morphological analysis are applied to each sentence.
Different from English, Japanese sentences were not explicitly split by such as white-space. According to previous researches, \cite{tokenizer2021,tokenizer2022,tokenizer2023} The application of morphological analysis can enhance the precision of model performance evaluation in comparison to the absence of such an analysis in the context of Japanese text. The difference between morphological analyzers hardly affects by performance. Therefore, MeCab \cite{mecab} is used as morphological analyzer. Although custom dictionaries, which include latest words such as neologd, is not necessary to improve the model performance, custom dictionary were made referred to J-Medic \cite{jmedic} in order to specialize clinical and medicine words.

In tokenization, Unigram tokenizer \cite{unigram} is used as tokenizer. Morphological analyzed text are split by white-space at pre-tokenizer. The vocabulary size was decided to be 32768 ($=2^{15}$) for NCVC-slm-1 base model. Special tokens are following: <|begin\_of\_text|> (means begin of text), <|end\_of\_text|> (means end of text and uses as padding tokens), and $\backslash$n (means new line). The hyperparameters of training Unigram tokenizer are set following: shrinking factors is 0.75, max piece length is 16, and number of sub-iterations is 8. The corpus for training tokenizer used Japanese Wikipedia and medicine textbooks (including both scraping and synthesized) in order to prioritize clinical and medicine words.

\subsection{Modeling}
\begin{figure}[htbp]
  \centering
  \includegraphics[scale=0.35]{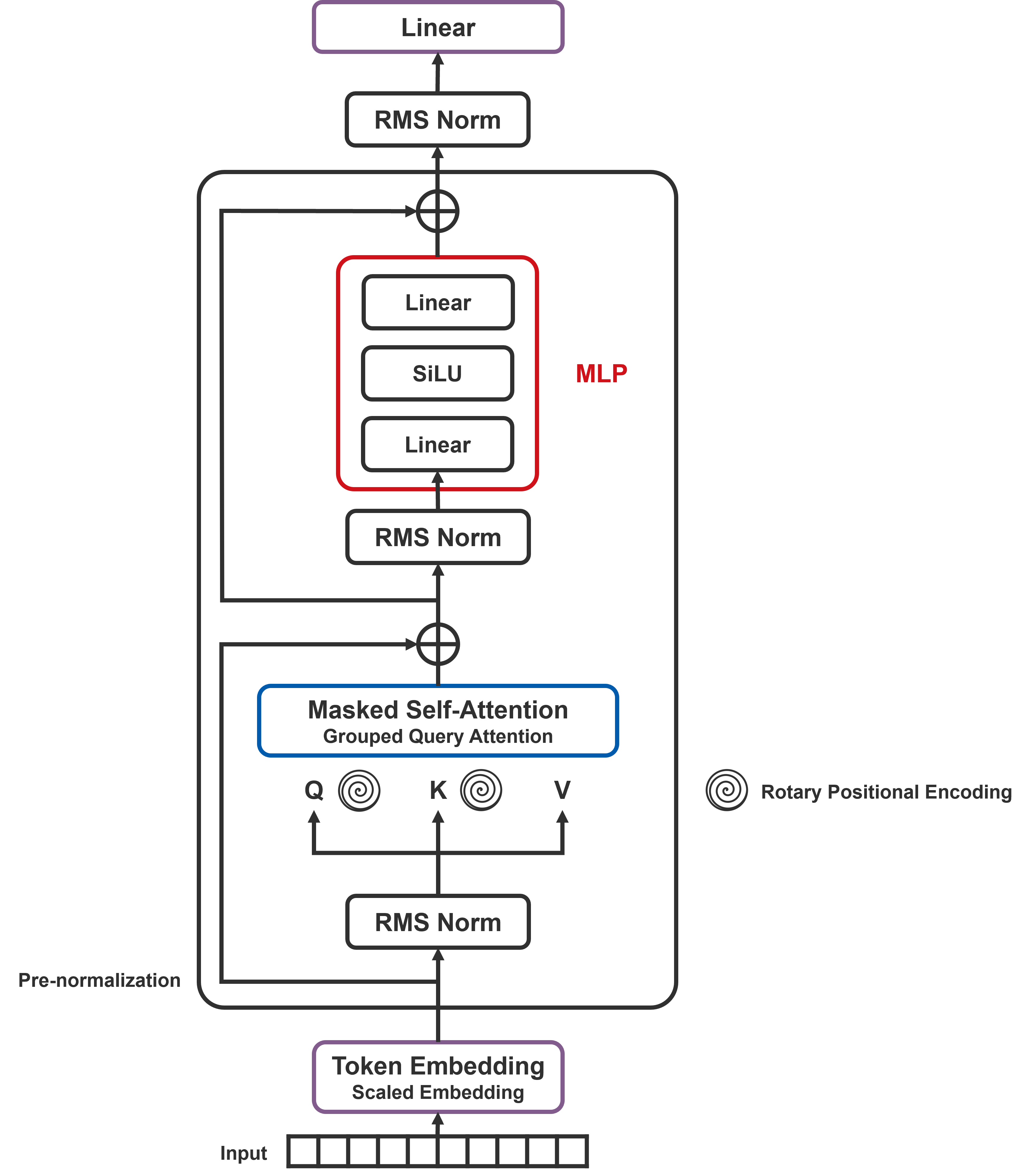}
  \caption{The overview of NCVC-slm-1 model architecture}
  \label{fig:model-architecutre}
\end{figure}

Figure \ref{fig:model-architecutre} illustrates the overview of NCVC-slm-1 architecture. 

NCVC-slm-1 is based on phi-1, LLaMA series, and the advantages of previous language models. Basic hyperparameters are following: 24 layers, 2048 hidden dimensions, 8192 internal hidden dimensions in MLP, 64 head dimensions, and 32 heads. The maximum sequence length is 2048 at pre-training.
Positional encoding (PE) plays an important role to Transformer-based model because of not considering the order of inputs explicitly. Rotary positional encoding \cite{rope2021} is one of the better approaches for PE. Similarly to several language models, NCVC-slm-1 used rotary PE in every query and key of self-attention layers.
A normalization architecture is applied pre-normalization because of stabilizing model training \cite{prenorm}.
The total number of parameters are approximately 1 billions (more accurately 1.2B), and the reserved GPU memory is only 2.2 GB memory with bfloat16 \cite{bfloat16} precision.

NCVC-slm-1 is modified and updated on several components from phi-1.

The activation function is replaced gaussian error linear units (GELU) (gelu\_new) \cite{gelu2016} with sigmoid linear units (SiLU) \cite{silu2017}. Although GELU has the features of differentiable at near 0 and similar effects to dropout and it has been used instead of ReLU \cite{relu} since around 2018, approximal GELU implementation requires the high computation burden. In preliminary experiments, 28\% GPU memory reduction were observed by replacing GELU (gelu\_new) with SiLU. In phi-3, SiLU also used as non-linear activation function instead of GELU (gelu\_new).

Layer normalization (LayerNorm) \cite{layernorm} is replaced with root mean square normalization (RMSNorm) \cite{rmsnorm} inspired by LLaMA series. RMSNorm reduces the computation burden comparing with LayerNorm. In actually, the series of phi-3 also used RMSNorm.

Multi-head attention (MHA) is replaced with grouped query attention (GQA) \cite{gqa} inspired by LLaMA series. Although MHA is replaced with multi-query attention (MQA) \cite{mqa} on the previous study \cite{falcon2023}, due to concerns that MQA is degraded the model performance, GQA was applied to NCVC-slm-1 in finally. The number of group is applied 8 refer to the proposed paper.

Initialization parameters are applied to several policies. Most parameters are initialized small init \cite{megatronlm2019,smallinit2019}. The hidden dimension of NCVC-slm-1 is 2048, therefore $\mathcal{N} (0, \sigma)$ $\left( \sigma = \sqrt{\frac{2}{5d}} = \sqrt{\frac{2}{5 \times 2048}} \approx 0.01 \right)$. Based on the previous research \cite{Takase2023} for stabilizing language model pre-training and preventing loss spike, scaled embedding \cite{Vaswani2017} for token embedding and scaled initialization approach is applied $W_O$ and $W_2$. Scaled initialization is derived by $\mathcal{N} \left( 0, \frac{\sigma}{\sqrt{2 N}} \right)$, where $N$ is number of hidden layers. In the initialization, very small standard deviation was used. $\left(N=24, 0.01 / \sqrt{2 \times 24} \approx 0.001 \right)$
The head of causal language model parameters is initialized Xavier normal \cite{glorot2010understanding}. Bias parameters are disabled in all layers.

\subsection{Training}

\subsubsection{Self-supervised Pre-training}
According to Chinchilla scaling law, 1B parameters model requires 20B tokens \cite{chinchilla2022}. The phi-1 paper reported total seen 50B tokens (of which 7B tokens were unique) for pre-training. On the other hand, excessive epoch learning, which repeatably seeing the same training data, declines language model performance \cite{nottorepeat}. Based on these previous findings, self-supervised pre-training was performed total 24000 global steps (save model per 3200 global steps) – this is equivalent to 5.5 epochs and total seen approximately 50B tokens (of which 9B tokens were unique).
Effective batch size were performed 1024 with flash attention 2 \cite{flashattn}, distributed data parallel, and DeepSpeed with zero redundant optimizer (ZeRO) stage 2 \cite{DeepSpeed,ZeRO}. Similar to other language model training, tokens are packed to a length of 2048 using concatenation or truncation for computational efficiency.
The batch size of 1024 is consisted of 8 minibatches per a GPU, 4 GPUs, and 32 gradient accumulation steps. 
Dropout \cite{dropout} was applied to all attention and residual layers with 0.1 in order to mitigate a generalization performance degradation using repeat tokens.
Loss function is cross entropy for next token prediction manner.
Optimizer algorithm was selected AdamW \cite{adamw} with $\beta_1, \beta_2$, and weight decay was 0.9, 0.95, and 0.1, respectively.
Learning rate scheduling was used WarmupCosineLR, warmup to $1.0 \times 10^{-3}$ by 750 global steps and then cosine decay to 0.
Self-supervised pre-training spent 14 days using 4 NVIDIA 6000 Ada 48GB memory.
 
\subsubsection{Fine-tuning}
Fine-tuning was performed based on instruction tuning \cite{instruct-tuning}. Instruction tuning dataset was used 8 JMED-LLM \cite{jmed-llm} dataset except test samples. The effective of with or without synthesized exercises dataset also was evaluated. Base models were used total seen 20B tokens (NCVC-slm-1-instruct-20B) and total seen 50B tokens (NCVC-slm-1-instruct-50B).
When instruction tuning is applied, additional 3 special tokens (<|system|>, <|user|>, and <|assistant|>) are added to the tokenizer and the dimension of token embedding layer is extended 32768 to 32771.
The batch size is 256 (8 minibatches per a GPU, single NVIDIA A6000 48GB memory, and 32 gradient accumulation steps) in JMED-LLM only fine-tuning. The batch size is 1024 (the same as pre-training) in fine-tuning with synthesized exercises dataset.
Loss function is cross entropy similar to self-supervised pre-training. However, the loss function is calculated only positions from immediately after <|assistant|> to <|end\_of\_text|>.
AdamW optimizer is used and $\beta_1, \beta_2$, and weight decay was 0.9, 0.95, and 0.01, respectively.
Learning rate scheduling was used WarmupCosineLR, warmup to $1.0 \times 10^{-4}$ by 50 global steps and then cosine decay to 0.
Instruction tuning was run 890 or 4850 global steps and spent 9 hours on JMED-LLM only or 21 hours on with synthesized exercises, respectively.
In order to enhance the model performance, noisy embeddings fine-tuning (NEFTune) \cite{neftune} were applied to hidden states after token embedding layer in training. The strength of noise was set to $\alpha=5$.

\subsection{Validation}
In this report, two benchmarks are used for performance evaluation.
IgakuQA \cite{igakuQA} is a benchmark test for Japanese national medical licensing examinations. IgakuQA contains 5 examinations in 2018--2022.
JMED-LLM \cite{jmed-llm} is proposed by LLM-JP to evaluate language model performance for clinical and medicine in the various aspects. JMED-LLM is consisted of 6 major tasks including question and answer (Q\&A), document classification, and named entity recognition (NER). JMMLU-Med is a Q\&A task with respect to biology and medicine extracted from JMMLU. CRADE is a classification task for the possibility of adverse events in case reports. RRTNM is a classification task for the stage of cancer refer to radiology reports of lung cancer patients. SMDIS is a classification task for detection of illness or symptoms from tweets. JCSTS is classification task for similarity of two sentences in case reports. MRNER-disease and MRNER-medicine is a NER for extracting symptoms or pharmaceuticals from medical reports, respectively. NRNER is a NER for extracting symptoms and pharmaceuticals from nursing records. JMMLU-Med, CRADE, RRTNM, SMDIS, and JCSTS are evaluated by Cohen's kappa \cite{CohenKappa} and accuracy. NER tasks are evaluated by partial or exact F1 score.


\section{Results}

\subsection{Self-supervised Pre-training}
\begin{figure}[htbp]
  \centering
  \includegraphics[scale=0.45]{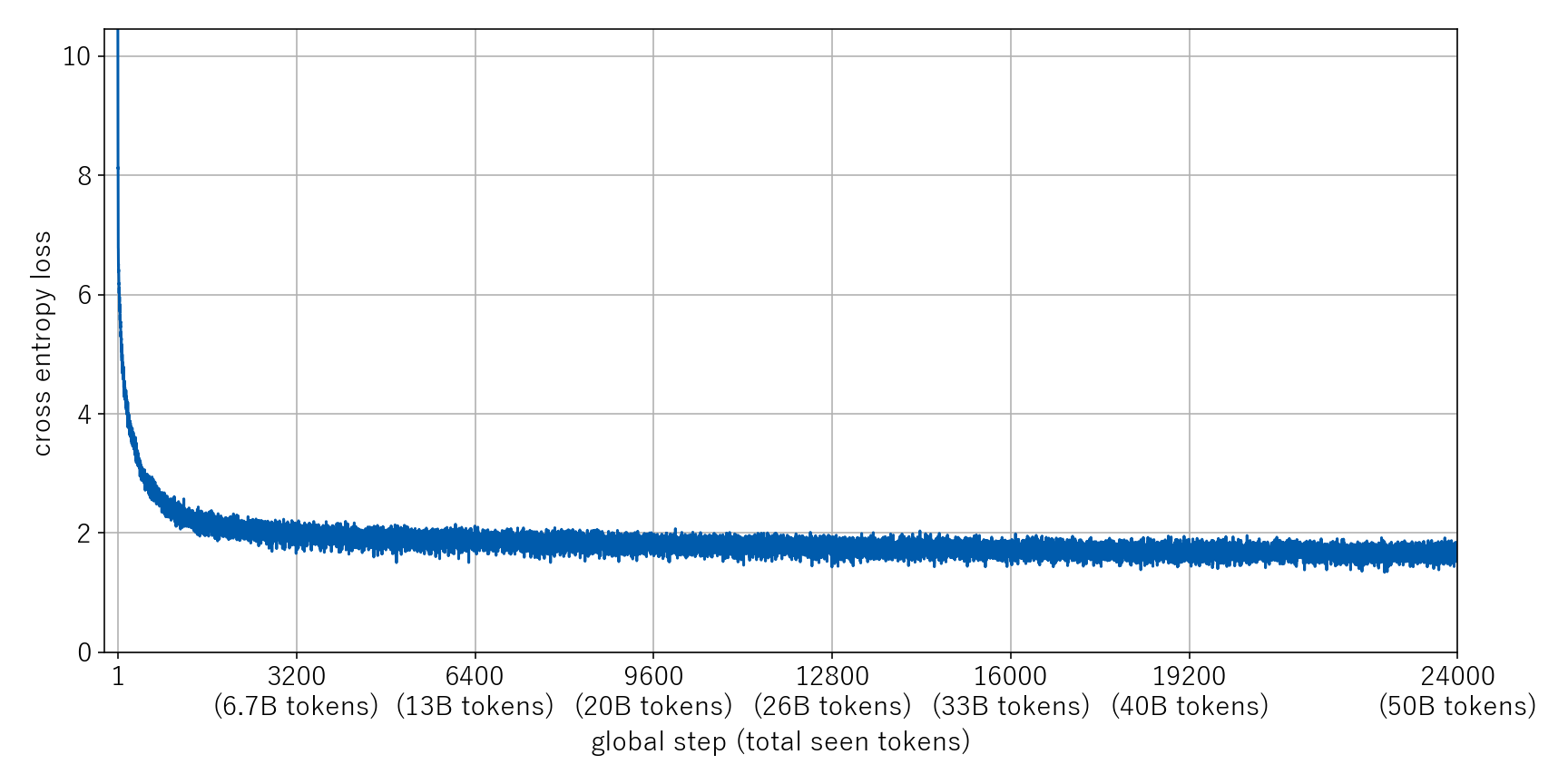}
  \caption{The loss logging during self-supervised pre-training}
  \label{fig:logging-pretraining}
\end{figure}

Figure \ref{fig:logging-pretraining} illustrates the logging of loss in self-supervised pre-training. The horizontal axis represents global steps and seen tokens. The vertical axis is global step and cross entropy loss, respectively. Loss function rapidly decreases until approximately 3200 global steps and slowly on subsequent steps. Despite the tendency towards stagnation, the value of loss function continues to decrease slightly. The loss spike, which destroy a model training, was not observed in this experiments.

\subsection{IgakuQA}
Table \ref{tab:igakuQA} shows the scoring board on IgakuQA. NCVC-slm-1-base and -instruct is displayed on the table. The scores of other LLMs were cited from the original article \cite{igakuQA} and the blog post \cite{pfnet-medswallow}. The instruct models are tuned by instructing only JMED-LLM dataset. The acceptance criteria is guessed approximately 75\%. The score of NCVC-slm-1 models remains below 20\%. There are no obvious difference between seen 20B and 50B tokens on both base and instruction. The scores of intermediate models saved per 3200 global steps are illustrated in Appendix \ref{apx:igakuqa-base-models}.

\begin{table}[htbp]
  \centering
  \caption{IgakuQA scores}
  \label{tab:igakuQA}
  \begin{tabular}{l|c|c|c|c|c}
   & 2018 & 2019 & 2020 & 2021 & 2022 \\ \hline
  GPT-4 & \textbf{382 (76.6\%)} & \textbf{385 (77.6\%)} & \textbf{387 (78.0\%)} & \textbf{398 (79.6\%)} & \textbf{392 (79.4\%)} \\ \hline
  ChatGPT (gpt-3.5-turbo) & 266 (53.3\%) & 250 (50.4\%) & 266 (53.6\%) & 297 (59.4\%) & 287 (58.1\%) \\ \hline
  GPT-3 (text-davinci-003) & 209 (41.9\%) & 210 (42.3\%) & 208 (41.9\%) & 203 (40.6\%) & 217 (43.9\%) \\ \hline
  c4ai-command-r-plus & 321 (64.3\%) & 303 (61.1\%) & 320 (64.5\%) & 302 (60.4\%) & 335 (67.8\%) \\ \hline
  Qwen2-72B & 320 (64.1\%) & 325 (65.5\%) & 325 (65.5\%) & 326 (65.2\%) & 360 (72.9\%) \\ \hline
  Llama3-Preferred-MedSwallow-70B & \textbf{407 (81.6\%)} & \textbf{390 (78.6\%)} & \textbf{391 (78.8\%)} & \textbf{393 (78.6\%)} & \textbf{395 (80.0\%)} \\ \hline
  Llama3-Swallow-70B-v0.1 & 353 (70.7\%) & 347 (70.0\%) & 353 (71.2\%) & 345 (69.0\%) & 345 (69.8\%) \\ \hline
  Meta-Llama-3-70B & 353 (70.7\%) & 340 (68.5\%) & 348 (70.2\%) & 314 (62.8\%) & 318 (64.4\%) \\ \hline
  Swallow-70b-NVE-hf & 283 (56.7\%) & 280 (56.5\%) & 300 (60.5\%) & 295 (59.0\%) & 300 (60.7\%) \\ \hline
  Swallow-MX-8x7b-NVE-v0.1 & 269 (53.9\%) & 285 (57.5\%) & 290 (58.5\%) & 277 (55.4\%) & 290 (58.7\%) \\ \hline
  gemma-2-27b & 337 (67.5\%) & 298 (60.1\%) & 327 (65.9\%) & 296 (59.2\%) & 322 (65.2\%) \\ \hline
  NCVC-slm-1-base (seen 20B tokens) & 70 (14.0\%) & 64 (12.9\%) & 73 (14.7\%) & 77 (15.4\%) & 57 (11.5\%) \\ \hline
  NCVC-slm-1-base (seen 50B tokens) & 62 (12.4\%) & 64 (12.9\%) & 83 (16.7\%) & 78 (15.6\%) & 56 (11.3\%) \\ \hline
  NCVC-slm-1-instruct (seen 20B tokens) & 102 (20.4\%) & 93 (18.8\%) & 77 (15.5\%) & 106 (21.2\%) & 95 (19.2\%) \\ \hline
  NCVC-slm-1-instruct (seen 50B tokens) & 115 (23.0\%) & 87 (17.5\%) & 84 (16.9\%) & 82 (16.4\%) & 80 (16.2\%) \\ \hline
  Total number of problems & 499 & 496 & 496 & 500 & 494
  \end{tabular}
\end{table}

\subsection{JMED-LLM}
Table \ref{tab:jmed-llm-QAC} and \ref{tab:jmed-llm-NER} show the scoring board on JMED-LLM. Similar to IgakuQA, base and instruct is displayed on the table. These models are the same as on IgakuQA. The scores of other LLMs were cited from \cite{jmed-llm-bench}. NCVC-slm-1 base models did not cope with the problems at all and consequently achieved the lowest scores in most tasks. On the other hand, instruction tuning models were the highest score of 6 tasks (CRADE, SMDIS, JCSTS, MRNER-disease, MRNER-disease, and NRNER). Nevertheless, the score of JMMLU-Med and RRTNM were comparable or below even if base models are enough fine-tuning. GPT-4o demonstrates the highest performance on these two tasks and is superior to other models.

\begin{table}[htbp]
  \centering
  \caption{JMED-LLM scores with Kappa(Accuracy)}
  \label{tab:jmed-llm-QAC}
  \begin{tabular}{l|c|c|c|c|c|c|c|c|c}
   & JMMLU-Med & CRADE & RRTNM & SMDIS & JCSTS \\ \hline
  GPT-4o-2024-08-06                 & \textbf{0.82(0.87)} & 0.54(0.53) & \textbf{0.85(0.90)} & 0.76(0.88) & 0.60(0.48) \\ \hline
  GPT-4o-mini-2024-07-18            & 0.77(0.83) & 0.21(0.37) & 0.58(0.71) & 0.56(0.78) & 0.57(0.51) \\ \hline
  gemma-2-9b-it                     & 0.52(0.64) & 0.33(0.42) & 0.54(0.68) & 0.62(0.81) & 0.16(0.24) \\ \hline
  Llama-3-ELYZA-JP-8B               & 0.34(0.51) & 0.01(0.26) & 0.29(0.52) & 0.54(0.77) & 0.02(0.18) \\ \hline
  Meta-Llama-3.1-8B-Instruct        & 0.31(0.49) & 0.11(0.32) & 0.41(0.57) & 0.28(0.64) & 0.13(0.23) \\ \hline
  Meta-Llama-3-8B-Instruct          & 0.42(0.57) & 0.00(0.25) & 0.37(0.54) & 0.43(0.72) & 0.16(0.24) \\ \hline
  Llama-3-Swallow-8B-Instruct-v0.1  & 0.33(0.50) & 0.31(0.37) & 0.33(0.55) & 0.26(0.63) & 0.01(0.17) \\ \hline
  Qwen2-7B-Instruct                 & 0.42(0.57) & 0.11(0.29) & 0.31(0.53) & 0.33(0.67) & 0.37(0.31) \\ \hline
  gemma-2-2b-it                     & 0.17(0.38) & 0.00(0.25) & 0.24(0.42) & 0.14(0.57) & 0.12(0.21) \\ \hline
  Llama-3-youko-8b-insturct         & 0.31(0.49) & 0.02(0.28) & 0.28(0.47) & 0.50(0.75) & 0.01(0.20) \\ \hline
  NCVC-slm-1-base (seen 20B tokens) & 0.03(0.27) & -0.00(0.16) & -0.07(0.10) & 0.01(0.01) & 0.05(0.18) \\ \hline
  NCVC-slm-1-base (seen 50B tokens) & 0.13(0.35) & -0.02(0.21) & -0.10(0.06) & 0.02(0.09) & -0.03(0.12) \\ \hline
  NCVC-slm-1-instruct (seen 20B tokens) & 0.19(0.39) & \textbf{0.61(0.67)} & 0.36(0.53) & \textbf{0.98(0.99)} & \textbf{0.75(0.67)} \\ \hline
  NCVC-slm-1-instruct (seen 50B tokens) & 0.26(0.44) & 0.50(0.59) & 0.42(0.58) & \textbf{0.98(0.99)} & \textbf{0.76(0.66)}
  \end{tabular}
\end{table}

\begin{table}[htbp]
  \centering
  \caption{NER scores with Partial F1(Exact F1)}
  \label{tab:jmed-llm-NER}
  \begin{tabular}{l|c|c|c|c|c|c|c|c|c}
   & MRNER-disease & MRNER-medicine & NRNER \\ \hline
  GPT-4o-2024-08-06                 & 0.54(0.15) & 0.42(0.26) & 0.39(0.20) \\ \hline
  GPT-4o-mini-2024-07-18            & 0.48(0.13) & 0.52(0.32) & 0.48(0.25) \\ \hline
  gemma-2-9b-it                     & 0.61(0.16) & 0.65(0.42) & 0.53(0.30) \\ \hline
  Llama-3-ELYZA-JP-8B               & 0.83(0.31) & 0.51(0.31) & 0.47(0.26) \\ \hline
  Meta-Llama-3.1-8B-Instruct        & 0.82(0.30) & 0.54(0.32) & 0.36(0.18) \\ \hline
  Meta-Llama-3-8B-Instruct          & 0.60(0.20) & 0.44(0.25) & 0.41(0.21) \\ \hline
  Llama-3-Swallow-8B-Instruct-v0.1  & 0.56(0.17) & 0.37(0.21) & 0.42(0.24) \\ \hline
  Qwen2-7B-Instruct                 & 0.24(0.06) & 0.29(0.14) & 0.33(0.17) \\ \hline
  gemma-2-2b-it                     & 0.66(0.20) & 0.46(0.23) & 0.46(0.26) \\ \hline
  Llama-3-youko-8b-insturct         & 0.02(0.00) & 0.05(0.02) & 0.11(0.07) \\ \hline
  NCVC-slm-1-base (seen 20B tokens) & 0.00(0.00) & 0.00(0.00) & 0.00(0.00) \\ \hline
  NCVC-slm-1-base (seen 50B tokens) & 0.00(0.00) & 0.00(0.00) & 0.00(0.00) \\ \hline
  NCVC-slm-1-instruct-20B           & \textbf{0.87(0.35)} & \textbf{0.65(0.38)} & \textbf{0.86(0.61)} \\ \hline
  NCVC-slm-1-instruct-50B           & \textbf{0.90(0.24)} & 0.55(0.34) & \textbf{0.88(0.66)}
  \end{tabular}
\end{table}

Table \ref{tab:igakuqa-it}, \ref{tab:jmed-llm-QAC-it}, and \ref{tab:jmed-llm-NER-it} shows the comparison of with or without synthesized exercises dataset for instruction tuning. In IgakuQA benchmark, the performance of instruction models was degraded by using synthesized exercises dataset. In the worst case (base-seen-20B with synthetic), the performance was worsen than base model.

\begin{table}[htbp]
  \centering
  \caption{The comparison of using synthesized exercises dataset or not (IgakuQA)}
  \label{tab:igakuqa-it}
  \begin{tabular}{l|c|c|c|c|c}
   & 2018 & 2019 & 2020 & 2021 & 2022 \\ \hline
  base-seen-20B (JMED-LLM only)  & 102 (20.4\%) & 93 (18.8\%) & 77 (15.5\%) & 106 (21.2\%) & 95 (19.2\%) \\ \hline
  base-seen-50B (JMED-LLM only)  & 115 (23.0\%) & 87 (17.5\%) & 84 (16.9\%) & 82 (16.4\%) & 80 (16.2\%) \\ \hline
  base-seen-20B (with Synthetic) & 42 (8.4\%)   & 44 (8.9\%)  & 30 (6.0\%)  & 25 (5.0\%)  & 33 (6.7\%) \\ \hline
  base-seen-50B (with Synthetic) & 78 (15.6\%)  & 73 (14.7\%) & 78 (15.7\%) & 57 (11.4\%) & 57 (11.5\%)
  \end{tabular}
\end{table}

In JMED-LLM benchmark, there are minor differences between tasks, but no obvious differences are appeared. However the score of JMMLU-Med is tend to be improved using synthesized exercises dataset.

\begin{table}[htbp]
  \centering
  \caption{The comparison of using synthesized exercises dataset or not (JMED-LLM scores) with Kappa(Accuracy)}
  \label{tab:jmed-llm-QAC-it}
  \begin{tabular}{l|c|c|c|c|c|c|c|c|c}
   & JMMLU-Med & CRADE & RRTNM & SMDIS & JCSTS \\ \hline
  base-seen-20B (JMED-LLM only) & 0.19(0.39) & 0.61(0.67) & 0.36(0.53) & 0.98(0.99) & 0.75(0.67) \\ \hline
  base-seen-50B (JMED-LLM only) & 0.26(0.44) & 0.50(0.59) & 0.42(0.58) & 0.98(0.99) & 0.76(0.66) \\ \hline
  base-seen-20B (with Synthetic) & 0.33(0.50) & 0.53(0.59) & 0.31(0.51) & 0.96(0.98) & 0.67(0.52) \\ \hline
  base-seen-50B (with Synthetic) & 0.32(0.49) & 0.51(0.60) & 0.41(0.57) & 0.92(0.96) & 0.70(0.63)
  \end{tabular}
\end{table}

\begin{table}[htbp]
  \centering
  \caption{The comparison of using synthesized exercises dataset or not (NER scores) with Partial F1(Exact F1)}
  \label{tab:jmed-llm-NER-it}
  \begin{tabular}{l|c|c|c|c|c|c|c|c|c}
   & MRNER-disease & MRNER-medicine & NRNER \\ \hline
  base-seen-20B (JMED-LLM only) & 0.87(0.35) & 0.65(0.38) & 0.86(0.61) \\ \hline
  base-seen-50B (JMED-LLM only) & 0.90(0.34) & 0.55(0.34) & 0.88(0.66) \\ \hline
  base-seen-20B (with Synthetic) & 0.91(0.31) & 0.62(0.34) & 0.92(0.63) \\ \hline
  base-seen-50B (with Synthetic) & 0.98(0.35) & 0.69(0.42) & 0.90(0.64)
  \end{tabular}
\end{table}

\subsection{Ablation Study}
In order to inspect the base model, it was performed to visualized token embedding space, to demonstrate token analogy, and to visualize attention map in packing tokens. NCVC-slm-1-base seen 20B tokens was used as the base model for ablation study. Considering the arrangement of pages, these figures are posted to Appendix. See more details on Appendix \ref{apx:embed-space}, \ref{apx:analogy}, and \ref{apx:attn-map}.


\section{Discussion}
Transformer-base LLMs have achieved grateful successes along with Scaling laws and drastically developed. However, it is said that the rate of performance improvement stagnates in recent years. Furthermore, it rises up the problem of the shortage of high-quality data for training contrary to the size of model parameters. From various perspectives, which include the limit of scaling up the size of model or energy problems, it may be to require smaller and smarter language models optimized the specific fields and particular tasks in the next stages. It is hoped that this report will contribute to the development of SLMs in conjunction with other previous studies.

It is considered that the development of a smart SLM require high-quality dataset. Based on this intuition, strong screening and filtering was applied to pre-training corpus in this project. Unlike an expectation, Transformer-based classifier of text quality was hardly filtering. There are several possible reasons for this result. One is that corpus was already enough filtering because 94\% of randomly sampled 100,000 texts were labeled to high-quality. Another one is that the instruction prompt for annotation was not optimized. In general, annotation is a very tough task. Moreover, it is too difficult to define how good text is. Although text quality annotations were automated by LLM to reduce the burden of annotation and to avoid the subjective evaluation, the criteria for classifying text quality were also unclear and LLM sometimes did not follow the instruction (e.g. LLM answered not in Japanese but in English or generated non-relation text etc.). It is not easy to operate LLMs by only instruction prompt. If you automate to classify text quality, it may be needed to develop an optimized annotation model. Furthermore, the computation costs were not small to derive the hidden vector. Considering benefits and efforts, it may be not effective to use this quality filtering.

In order to augment the clinical and medicine contents, synthesized textbooks approach were also performed following the phi-1 paper. In this project, only 0.23B tokens were prepared despite considering variation of models and topics. This is very fewer than total tokens of corpus. It is necessary to add much more clinical and medicine knowledge and to train various documents such as radiological, nursing, and summary report.
Synthesized approach has been focused on LLM training recently. Considering the problem of training data shortage, this approach will be utilized more and more. On the other hand, synthesized approach contains the problem of curse of recursion \cite{CurseOfRecursion}. Therefore, it must be concerned with the proportion of real and synthesized data. Well, where is more high-quality text? Valuable text may be buried in traditional analog textbooks. In the current stage, it cannot be to use the contents of these textbooks for model training. But, it will be necessary to cooperate with us for a bright future.

In other language models which mainly developed in English, a lot of Japanese words are split to character-level or decomposed by fallback to UTF-8. Character- or Unicode-level tokenization is effective to avoid unknown tokens, but it is possible to be lost meanings contained in words or abstract concepts represented in words. These disadvantages may be affected to understanding the sentences and documents in the specific field. NCVC-slm-1 uses morphological analysis before tokenization. It is considered that this pre-processing is contributed to acquire a refined token embedding space.

In self-supervised pre-training, some methods and techniques were combined in order to prevent from encountering loss spike. Loss spike was not observed in the final experiment. However, it was sometimes observed in preliminary trials. In particular, it appeared in the initial steps on training. Although a loss spike was observed in the intermediate steps at once, pre-training was not collapsed. Therefore, it is needed to carefully observe the loss deviation in the early stage of pre-training. However, further investigation may be required to larger models.

IgakuQA is a benchmark with respect to clinical and medicine knowledge. A refined pre-trained language model is able to answer using few- or zero-shot learning. In pre-training of this project, synthesized exercises dataset was included in the pre-training dataset for the purpose of enhancing the ability of base model to solve the problems. However, its effect was not enough to be expected. In addition, the amount of pre-training tokens was not affected by the improvement of base model performance between seen 20B and 50B tokens. This result is consistent with Chinchilla scaling law. In contrast, recent other language models, such as Meta LLaMA, Google gemma, etc., use many more tokens for pre-training than the number of tokens estimated by that law, and the model performance is improved continuously. The relationship between the size of model parameters and the number of training tokens is unclear. Still, it guesses that the more training tokens increase, the better a base model performance improves. It is possible that limited unique tokens and less clinical and medicine content caused the stagnation of improvement in this project. The instruct models did not improve but declined the scores by adding synthesized exercises dataset. Data augmentation improves the model performance intuitively. However, it is possible that the quality of synthesized dataset was low. It suggests that the quantity and quality of fine-tuning dataset are required to enhance model performance.

JMED-LLM contains comprehensive tasks related to language processing in the field of clinical and medicine. The benchmark presents various instructions for each task. A base model is just an auto-regressive language model; therefore, it can barely answer a simple form Q\&A using few-shot learning manner, but it does not understand complex instruction and what it should do for other tasks without any fine-tuning (even if it has rich knowledge). It is possible that a base model can no more answer complex tasks than human beings who did not practice solving the exercises can (except extremely large models seen numerous tokens in pre-training). Instruction tuning is one of the better fine-tuning methods to improve language model output quality and conformity, and this method is widely used in developing other LLMs. NCVC-slm-1-instruct was fine-tuned using this method, and an instruct model achieved high scores on 6 tasks. It suggests that SLM is an effective choice to perform particular tasks by focusing on the domain. In the case of clinical settings, it is desirable for SLM to process some tasks, which include personal and sensitive information in completely local environments. However, it did not improve the performance of JMMLU-Med and RRTNM less than other tasks. It is considered that these two tasks require a high inference ability, and it is guessed that the limitation of 1B parameter models causes this result. The sophisticated inference ability, such as understanding the backgrounds and circumstances, extracting accurate knowledge from memories, and selecting how to respond appropriately, is required to solve complex problems. As one of the future strategies, it indicates that enough pre-trained and fine-tuned SLMs process some tasks quickly, while more huge LLMs are used for more complex and difficult tasks.

This study has several limitations.
First, a training corpus is limited to Web-based text. It did not include more expert clinical and medicine knowledge and the actual medical text, such as nursing and radiological reports or summaries. It is necessary to apply continuous pre-training or fine-tuning for deployment in real-world settings. It is considered that the volume of clinical and medicine content is inadequate.
Second, a training corpus contains synthesized dataset. These text data may cause the curse of recursion and the hallucinations. Hallucinations is inevitable problems when using LLMs. It inevitably needs a fact check.
Third, although pre-training corpus is supposed to have been strongly screening and filtering, it is possible that the model absorbed bias and noise in training corpus.
Finally, instruction tuning dataset was limited. That may cause an overfitting because of depending on training dataset and some tasks. In fact, the degradation of some instruct models was observed such as generating too short outputs. It is assumed that enough and various kinds of high-quality dataset are required in order to acquire generalized performance (even if it aims at specializing in a particular task).


\section{Conclusion}
A SLM specialized particular language, specific field, and well fine-tuning is capable of performing useful against particular tasks. This type of SLM is expected to to assist human jobs in local environment that isolated network connections or limited computational costs.


\section*{Acknowledgments}
This research and development was funded by Idorsia pharmaceuticals Japan Ltd. Hardware was supported by HPC TECH Corporation.

\bibliographystyle{unsrt}  
\bibliography{references}  


\appendix
\section{Appendix}

\subsection{Text Generation Samples}

\begin{itembox}[l]{Text Generation Samples}
脳卒中と循環器病について教えてください。\\
\\
循環器病は、脳梗塞・心筋梗塞が多く、高齢者の発症割合が高い病気です。また、循環器病は糖尿病を合併した割合が非常に高く、生活習慣病としては、高脂血症・高血圧症・糖尿病などがあります。循環器疾患の中でも「狭心症」「心筋梗塞」は比較的多く、「心房細動」や「冠攣縮性狭心症」といった心臓の病気はあまり知られていません。しかし、心不全は心疾患のうち最も頻度の高い病気です。
循環器病の治療方針は、まず、症状が軽い場合は、血圧を下げることに焦点を当てます。そして、症状が重い場合は、心臓の機能を回復させ、それにより合併症を抑えることに焦点を当てます。そして、心臓移植・冠動脈バイパス術・不整脈治療薬などの補助療法が積極的に用いられています。
\end{itembox}

\subsection{Token Embedding Space Visualization}
\label{apx:embed-space}
\begin{figure}[htbp]
  \centering
  \includegraphics[scale=0.35, angle=90]{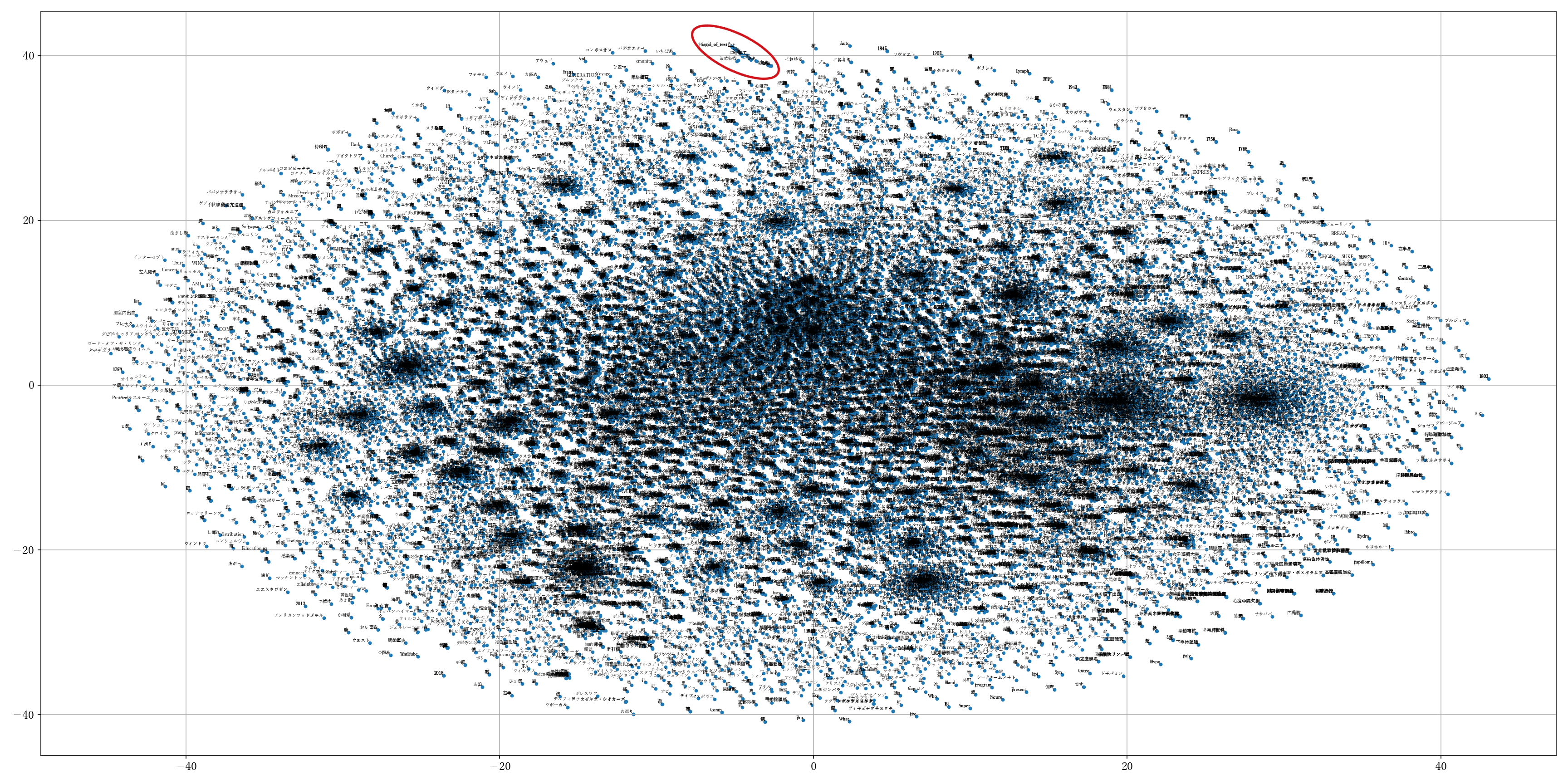}
  \caption{Token embedding space by t-SNE (perplexity=15)}
  \label{fig:tsne}
\end{figure}

\begin{figure}[htbp]
  \centering
  \includegraphics[scale=0.35, angle=90]{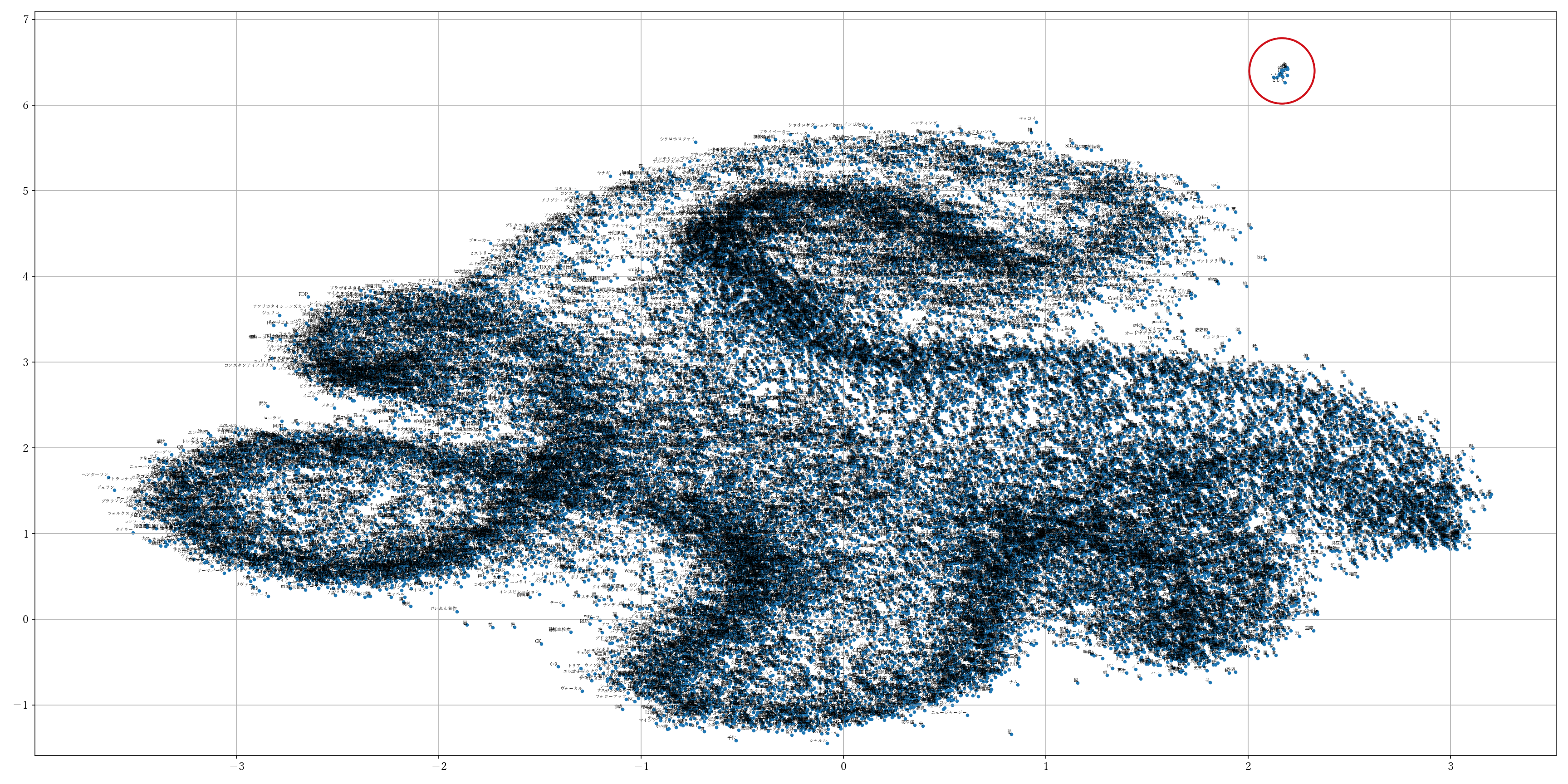}
  \caption{Token embedding space by UMAP (neighbor=15)}
  \label{fig:umap}
\end{figure}

Figure \ref{fig:tsne} and \ref{fig:umap} illustrate token embedding spaces using manifold learning algorithms and applying dimensional reduction for visualization. Figure \ref{fig:tsne} illustrates a token embedding space applied t-distributed Stochastic Neighbor Embedding (t-SNE) \cite{tSNE}. A hyperparameter perplexity was set to be 15. Figure \ref{fig:umap} illustrates a token embedding space applied Uniform Manifold Approximation and Projection (UMAP) \cite{umap}. A hyperparameter neighbor was set to be 15. The certain clusters has been observed in 2-dimensional space. Some stop-words or function words such as postpositional particle and special tokens are clustered at the red circles.

\subsection{Token Analogy}
\label{apx:analogy}
Word analogy was famous to Word2Vec \cite{Mikolov2013}. A well trained language model acquires the ability to analogy the relationships between words. It shows below some examples with respect to diseases.

\begin{itembox}[l]{Examples of token analogy}
脳出血 - 脳 + 消化管 = (cerebral hemorrhage - brain + gastrointestinal = ) \\
\quad 消化管出血 0.85 (gastrointestinal bleeding) \\
\quad 脳内出血 \quad 0.64 (intracerebral hemorrhage) \\
\quad 脳血管障害 0.63 (cerebrovascular disease) \\
\quad 血便 \quad\quad\quad 0.62 (hematochezia) \\
\quad 胃潰瘍 \quad\quad 0.61 (peptic ulcer) \\

尿路結石 - 尿路 + 胆管 = (urinary calculi - urinary tract + bile duct) \\
\quad 結石 \quad\quad 1.00 (calculus) \\
\quad 胆石 \quad\quad 0.97 (gallstone) \\
\quad 胆石症 \quad 0.96 (gallstone disease) \\
\quad 尿管結石 0.94 (ureteric stone) \\
\quad 胆管結石 0.92 (bile duct stones) \\

脳卒中 - 脳 + 心臓 = (stroke - brain + heart) \\
\quad 心疾患 \quad\quad\quad 0.99 (heart disease) \\
\quad 心不全 \quad\quad\quad 0.95 (heart failure) \\
\quad 脳梗塞 \quad\quad\quad 0.85 (ischemic stroke) \\
\quad 急性心筋梗塞 0.84 (acute myocardial infarction) \\
\quad 不整脈 \quad\quad\quad 0.82 (arrhythmia) \\

脳梗塞 - 脳 + 心臓 = (ischemic stroke - brain + heart) \\
\quad 心不全 \quad\quad\quad 1.00 (heart failure) \\
\quad 筋梗塞 \quad\quad\quad 1.00 (myocardial infarction) \\
\quad 梗塞 \quad\quad\quad\quad 1.00 (infarction) \\
\quad 心疾患 \quad\quad\quad 1.00 (heart disease) \\
\quad 急性心筋梗塞 0.96 (acute myocardial infarction) \\

大腸癌 - 大腸 + 肝臓 = (colorectal cancer - colon + liver) \\
\quad 肝細胞癌 1.00 (hepatocellular carcinoma) \\
\quad 肝癌 \quad\quad 1.00 (liver cancer) \\
\quad 肝転移 \quad 0.95 (liver metastases) \\
\quad 肝障害 \quad 0.92 (hepatic impairment) \\
\quad 胃癌 \quad\quad 0.88 (stomach cancer)
\end{itembox}

\newpage

\subsection{Base Model Performance}
\label{apx:igakuqa-base-models}
\begin{figure}[htbp]
  \centering
  \includegraphics[scale=0.45]{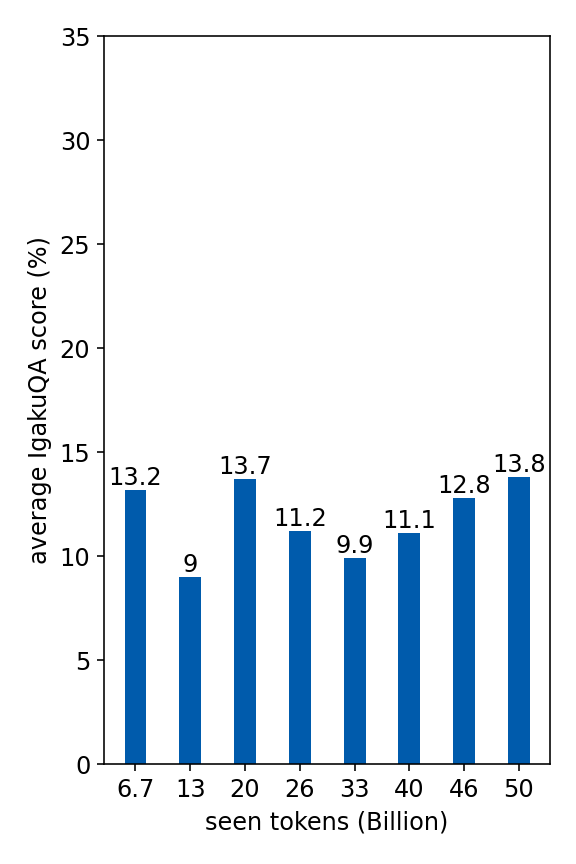}
  \caption{The relationship between seen tokens on pre-training and model performance}
  \label{fig:logging-IgakuQA}
\end{figure}

Figure \ref{fig:logging-IgakuQA} illustrates the comparison of between total seen tokens and average score on IgakuQA. It is observed the cycle of deterioration and improvement.

\subsection{Instruction Tuning}
\begin{figure}[htbp]
  \centering
  \includegraphics[scale=0.45]{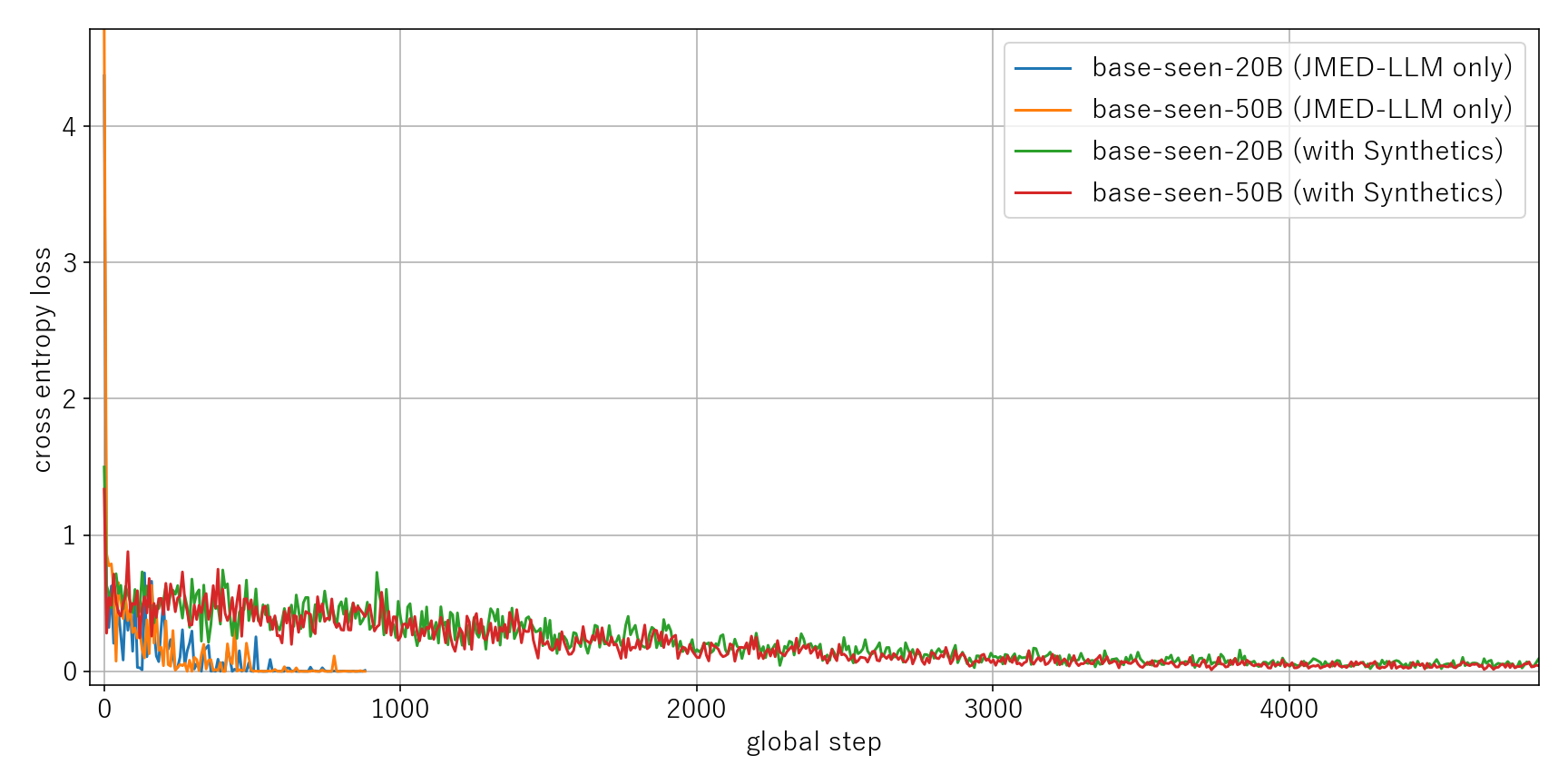}
  \caption{The loss logging of fine-tuning}
  \label{fig:logging-finetuning}
\end{figure}

Figure \ref{fig:logging-finetuning} illustrates the loss logging of instruction tuning. The horizontal and vertical axis is global step and cross entropy loss, respectively.

\subsection{Attention Map in Packing Tokens}
\label{apx:attn-map}
In pre-training, packing tokens are usually implemented for computation efficiency. The height and width of attention maps are query and key tokens. The horizontal ticks represent just sequence positions. The vertical ticks represent the position of <|end\_of\_text|> tokens. In causal language models, masking are applied and only lower triangular matrix are valid, therefore attention map scores are all zero on the upper right area. It is observed that the sequence of packed tokens are split into clusters by <|end\_of\_text|> in deeper and deeper layers.

\begin{figure}[htbp]
  \centering
  \includegraphics[scale=0.5]{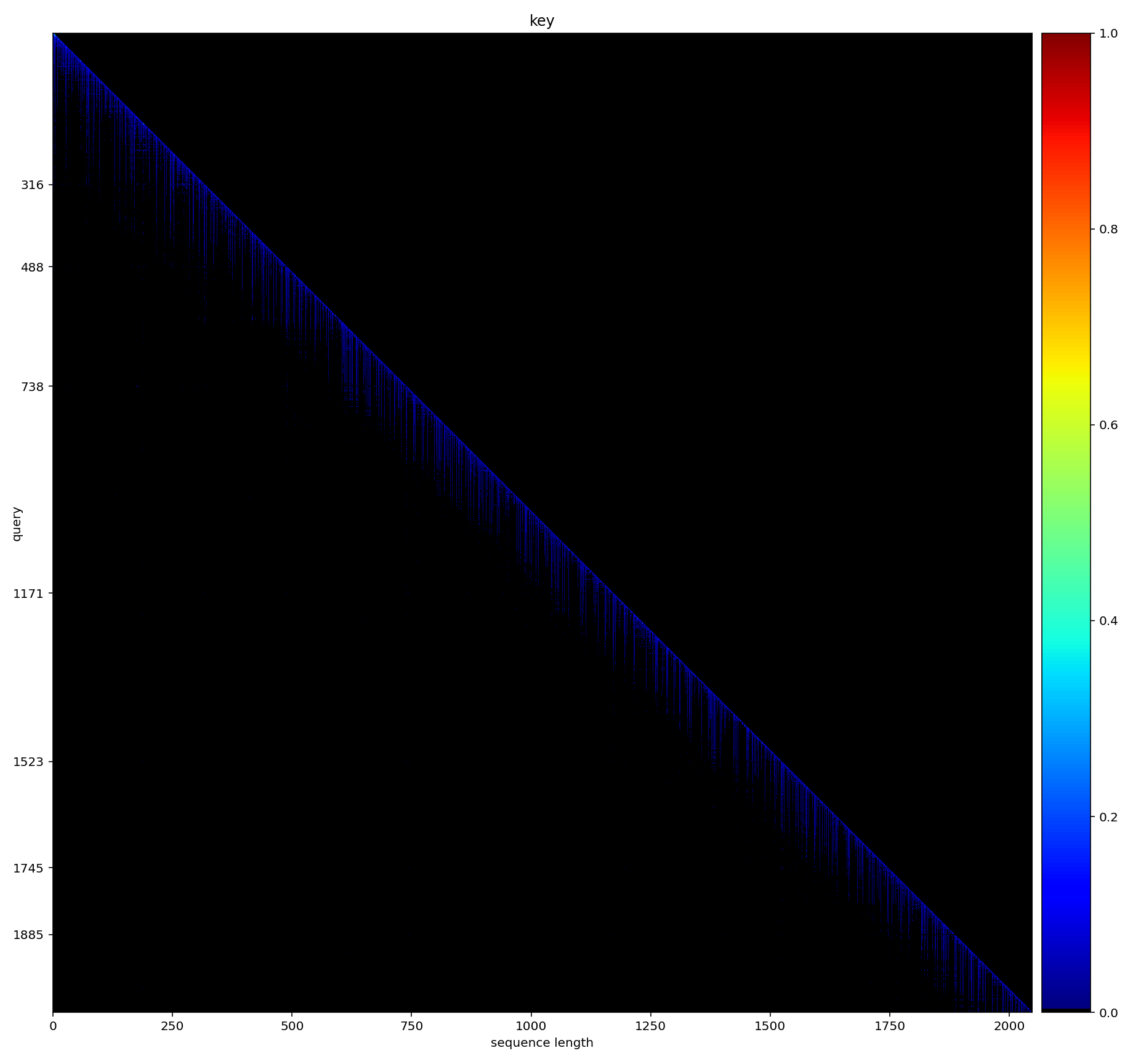}
  \caption{Attention map layer 1}
  \label{fig;layer1}
\end{figure}

\begin{figure}[htbp]
  \centering
  \includegraphics[scale=0.5]{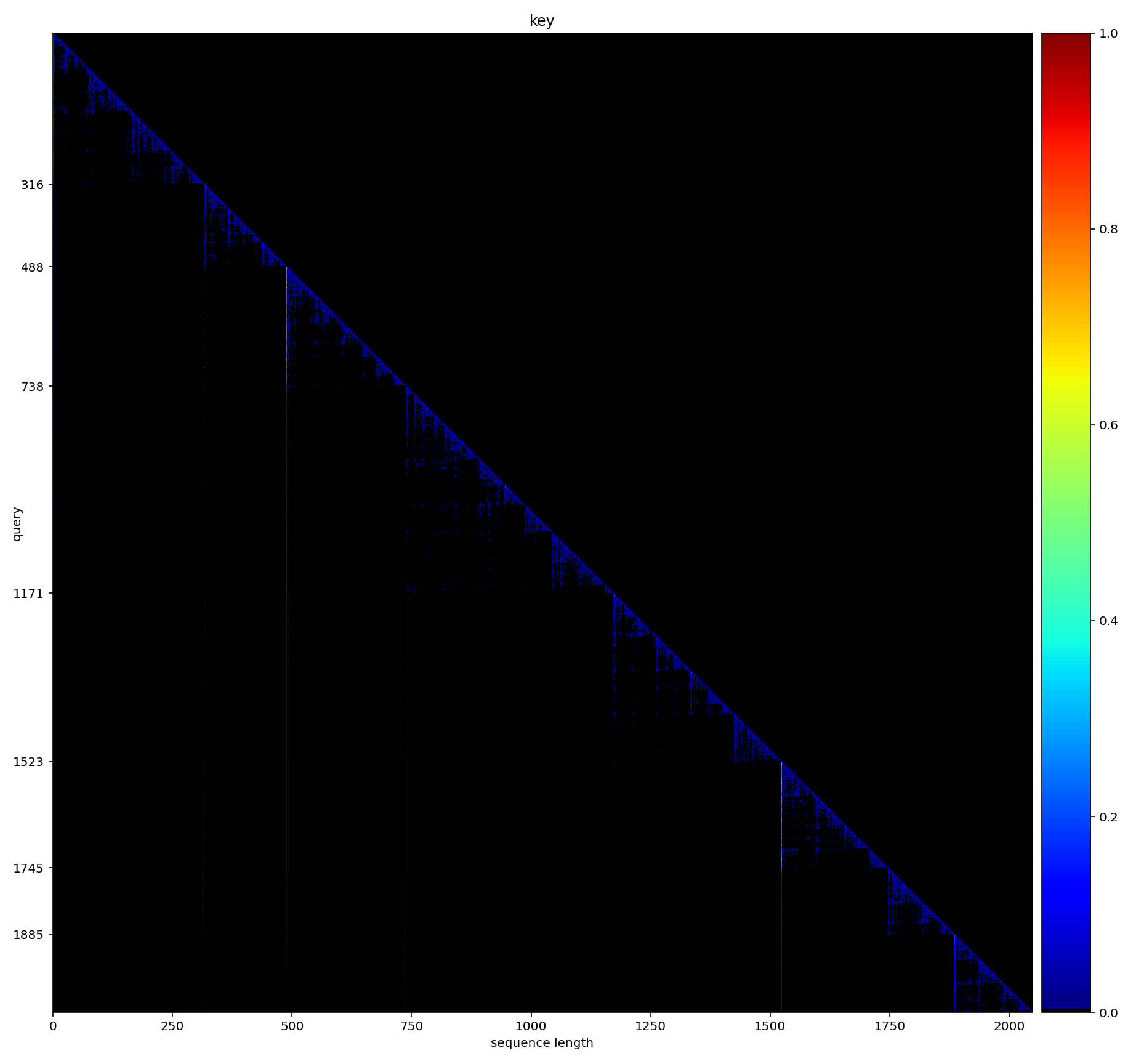}
  \caption{Attention map layer 12}
  \label{fig:layer12}
\end{figure}

\begin{figure}[htbp]
  \centering
  \includegraphics[scale=0.5]{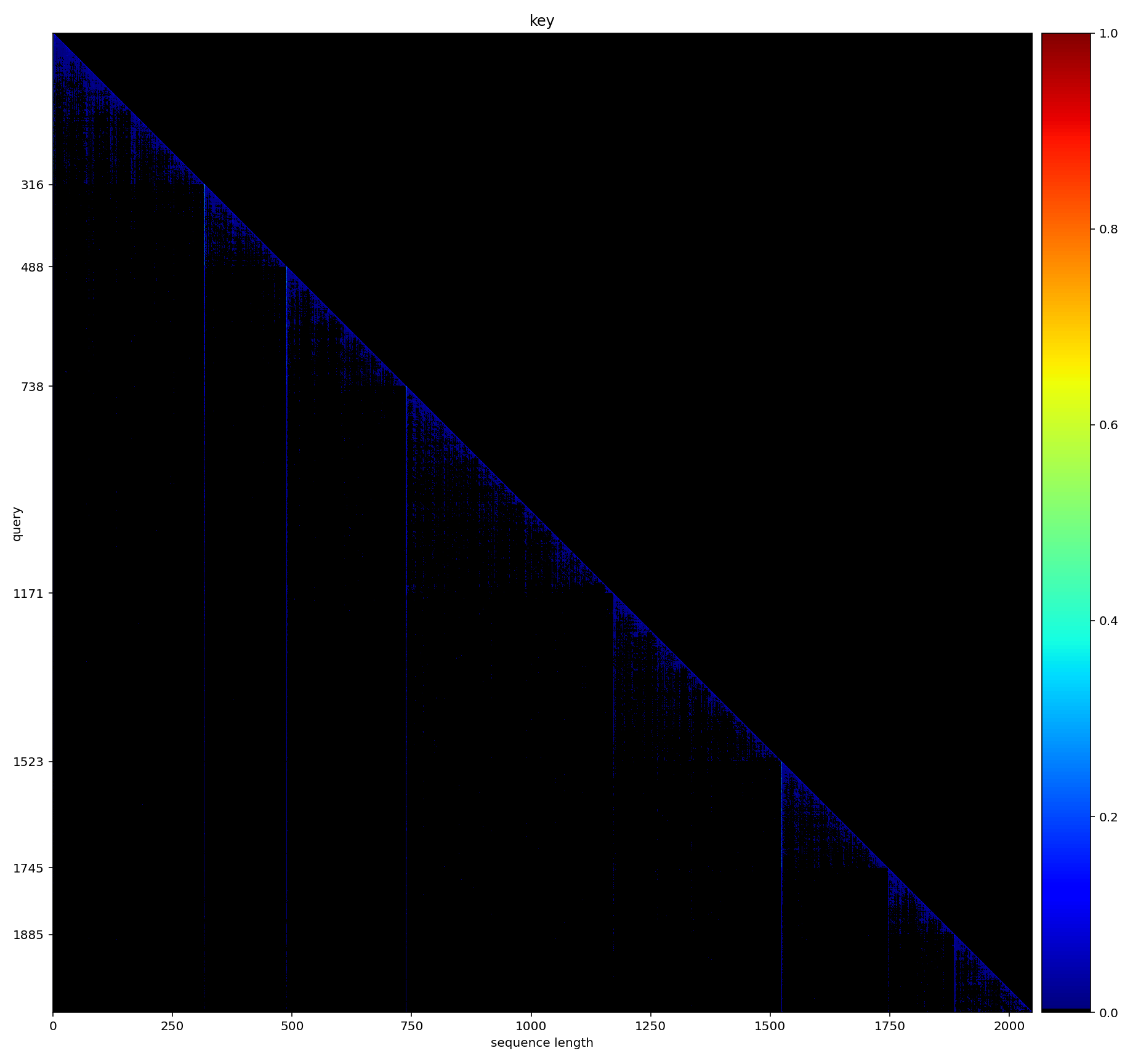}
  \caption{Attention map layer 24}
  \label{fig:layer24}
\end{figure}

\end{document}